\documentclass[acmsmall]{acmart}
\AtBeginDocument{%
  }

\setcopyright{none}

\usepackage{graphicx}
\usepackage{amsmath}
\usepackage{amssymb}
\usepackage{hyperref}
\usepackage{tikz}
\usepackage{booktabs}
\usepackage{caption}
\usepackage{array}
\usepackage{multirow}
\usepackage{amssymb} % for checkmark
\usepackage{pifont}  % for xmark
\usetikzlibrary{shapes.geometric, arrows, positioning}
\tikzstyle{process} = [rectangle, minimum width=2.5cm, minimum height=1.5cm, text centered, draw=black, fill=blue!20]
\tikzstyle{data} = [rectangle, minimum width=2cm, minimum height=1.2cm, text centered, draw=black, fill=green!20]
\tikzstyle{arrow} = [thick,->,>=stealth]

\begin{document}

\title{EfficientGFormer: Graph-Augmented Transformers for Multimodal Brain Tumor Segmentation}

%%
%% The "author" command and its associated commands are used to define
%% the authors and their affiliations.
%% Of note is the shared affiliation of the first two authors, and the
%% "authornote" and "authornotemark" commands
%% used to denote shared contribution to the research.
\author{Fatemeh Ziaeetabar}
\authornote{Corresponding author}
\email{fziaeetabar@ut.ac.ir}
\orcid{0000-0003-1159-3588}

\affiliation{%
  \institution{School of Mathematics, Statistics and Computer Science, College of Science, University of Tehran}
  \city{Tehran}
  %\state{Ohio}
  \country{Iran}
}

%\author{Florentin W\"org\"otter }
%\email{worgott@gwdg.de}
%\orcid{0000-0001-8206-9738}
%
%\affiliation{%
%  \institution{Bernstein Center for Computational Neuroscience, Department for Computational Neuroscience, III Physikalisches Institut-Biophysik, Georg-Agust-Universit\"at G\"ottingen, G\"ottingen}
%  \city{G\"ottingen}
%  %\state{Ohio}
%  \country{Germany}
%}

%%
%% The abstract is a short summary of the work to be presented in the
%% article.

\begin{abstract}
Accurate and efficient brain tumor segmentation remains a critical challenge in neuroimaging due to the heterogeneous nature of tumor subregions and the high computational cost of volumetric inference. In this paper, we propose \textbf{EfficientGFormer}, a novel architecture that integrates pretrained foundation models with graph-based reasoning and lightweight efficiency mechanisms for robust 3D brain tumor segmentation. Our framework leverages nnFormer as a modality-aware encoder, transforming multi-modal MRI volumes into patch-level embeddings. These features are structured into a \emph{dual-edge graph} that captures both spatial adjacency and semantic similarity. A pruned, edge-type-aware Graph Attention Network (GAT) enables efficient relational reasoning across tumor subregions, while a distillation module transfers knowledge from a full-capacity teacher to a compact student model for real-time deployment. Experiments on the MSD Task01 and BraTS 2021 datasets demonstrate that EfficientGFormer achieves state-of-the-art accuracy with significantly reduced memory and inference time, outperforming recent transformer-based and graph-based baselines. This work offers a clinically viable solution for fast and accurate volumetric tumor delineation, combining scalability, interpretability, and generalization.
\end{abstract}

\maketitle

\section{Introduction}

Accurate and efficient segmentation of brain tumors from multimodal magnetic resonance imaging (MRI) is a cornerstone task in neuro-oncology, directly informing diagnosis, prognosis, and treatment planning~\cite{menze2015multimodal}~\cite{bakas2017advancing}. The complexity of gliomas, characterized by heterogeneous subregions such as enhancing tumor (ET), tumor core (TC), and surrounding edema, poses persistent challenges due to overlapping intensity profiles, variable morphology, and inter-subject variability~\cite{isensee2021nnu}. Modern segmentation pipelines rely on supervised deep learning techniques that process 3D volumetric MRI data across four modalities (T1, T1ce, T2, FLAIR), each highlighting different tissue characteristics~\cite{bakas2017advancing}. Yet, despite substantial progress, achieving both high accuracy and computational efficiency remains elusive, especially in settings constrained by memory, latency, or clinical deployment demands.

Recent transformer-based models such as UNETR~\cite{hatamizadeh2022unetr}, Swin UNETR~\cite{hatamizadeh2022swin}, and nnFormer~\cite{zhou2022nnformer} have demonstrated remarkable performance on 3D segmentation tasks, benefiting from long-range context aggregation. However, their reliance on global attention over dense voxel representations leads to scalability bottlenecks in both training and inference. Moreover, these architectures predominantly treat volumetric MRI as raw tensor inputs, underutilizing the topological and structural priors that could guide more interpretable and sample-efficient learning.

In parallel, Graph Neural Networks (GNNs) have emerged as a compelling alternative for modeling relational structures in medical images~\cite{parisot2018disease}. By encoding spatial or appearance-based relationships as graph edges, GNNs offer a natural representation for capturing inter-region dependencies. Nevertheless, most prior GNN-based models either ignore modality heterogeneity or fail to incorporate fine-grained edge semantics, limiting their scalability and expressivity in 3D segmentation settings.

To address these gaps, we propose \textbf{EfficientGFormer}—a hybrid, structure-aware segmentation framework that combines the volumetric capacity of pretrained transformer encoders with the efficiency and interpretability of graph-based reasoning. Graph-based modeling and foundation models have shown promise across various domains, including video understanding ~\cite{ziaeetabar2024hierarchical}~\cite{ziaeetabar2025foundation}, motivating their integration for structured and multimodal reasoning in medical imaging.

\noindent \textbf{Our key contributions are summarized as follows:}
\begin{itemize}
  \item We propose \textbf{EfficientGFormer}, a novel graph-augmented transformer architecture for multi-label brain tumor segmentation, which integrates foundation model encoders, graph-based reasoning, and pruning-aware efficiency modules. 
  \item We introduce a \emph{dual-edge graph construction strategy} that explicitly captures both spatial contiguity and semantic similarity between image patches, enabling rich contextual modeling of tumor subregions.
  \item We design a \emph{pruned edge-type-aware Graph Attention Network (GAT)} that performs relational reasoning with minimal computational overhead, leveraging structured attention pruning to reduce inference cost.
  \item We implement a \emph{knowledge distillation module} that transfers the representational power of the full model to a lightweight student network, achieving real-time segmentation with negligible accuracy loss.
  \item We evaluate \textbf{EfficientGFormer} on the Medical Segmentation Decathlon (MSD) Brain Tumor dataset~\cite{antonelli2022medical} and the BraTS 2021 benchmark~\cite{baid2021rsna}, demonstrating state-of-the-art performance across multiple metrics (DSC, HD95) while achieving real-time inference speeds.
\end{itemize}

Compared to recent transformer-based and graph-based baselines, our model exhibits superior generalization, modular interpretability, and scalability. We believe this hybrid approach represents a promising direction for structure-aware, resource-efficient volumetric segmentation in real-world clinical systems.

\section{Related Works}

Accurate brain tumor segmentation from multimodal MRI has long been a central task in medical image analysis, with recent advances predominantly driven by deep learning. Below, we outline relevant efforts in volumetric segmentation using convolutional backbones, transformer-based encoders, graph neural networks, and hybrid architectures for efficient inference.

\textbf{Transformer-Based Volumetric Segmentation.}
Transformers have demonstrated remarkable capabilities in a range of computer vision tasks, particularly due to their inherent ability to model long-range dependencies and capture global contextual information. This strength has motivated the adaptation of transformer architectures to the domain of 3D medical image segmentation, where volumetric spatial relationships are essential for precise anatomical delineation.

Pioneering this trend, UNETR~\cite{hatamizadeh2022unetr} introduced a hybrid transformer architecture by integrating a ViT-based encoder with a U-Net-style decoder. By treating volumetric MRI data as a sequence of non-overlapping 3D patches and applying global self-attention, UNETR enabled holistic contextual understanding, outperforming conventional CNN-based models. Swin UNETR~\cite{hatamizadeh2022swin} improved upon this by employing Swin Transformers as encoders, which utilize a hierarchical representation and shifted-window attention mechanism to better capture multi-scale features while maintaining computational tractability.

Building on these foundations, nnFormer~\cite{zhou2022nnformer} incorporated domain-specific modifications tailored to medical imaging, including multi-scale patch tokenization and inter-patch attention fusion. This model demonstrated state-of-the-art performance across several 3D segmentation benchmarks, particularly on the Medical Segmentation Decathlon (MSD) tasks. It effectively reduced the attention overhead through localized window-based attention while preserving global semantic context.

Despite these advancements, transformer-based volumetric segmentation remains computationally intensive. The quadratic complexity of self-attention over a large number of 3D tokens results in high memory consumption and long inference times, posing significant challenges for deployment in real-time clinical workflows. Moreover, these models typically require extensive pretraining or large labeled datasets to generalize effectively, which may not always be available in medical imaging settings.

Our work builds upon these transformer-based innovations by incorporating a pretrained nnFormer encoder to extract high-quality, modality-specific features. We further mitigate the computational cost of downstream reasoning by combining transformer embeddings with a pruned, edge-aware graph structure, enabling efficient and scalable inference without compromising segmentation accuracy.

\textbf{Graph Neural Networks in Medical Imaging.}
Graph-based models have emerged as a powerful paradigm in medical image analysis, offering the ability to model complex relational structures beyond the capabilities of traditional convolutional architectures. In the context of segmentation, where anatomical continuity and inter-region dependencies are critical, graph neural networks (GNNs) provide a principled way to encode both spatial adjacency and semantic similarity.

Early efforts in GNN-based medical imaging primarily focused on population-level analysis or disease classification using handcrafted graph structures~\cite{parisot2018disease}. While effective for global representation learning, these methods were limited in their applicability to dense voxel-wise tasks. Subsequent approaches aimed to apply GNNs at the patch or superpixel level for segmentation, often using basic graph convolutional layers to propagate features among neighboring regions. However, such approaches typically lacked the capacity to model higher-order contextual dependencies or integrate multiple types of relational information.

Recent models like DGRUnit~\cite{liu2023dgrunit} introduced a dual-graph design, separating spatial and contextual relations into parallel reasoning streams. While this improved the representation of inter-region dependencies, the architecture did not incorporate edge-type awareness or feature-adaptive pruning, and remained computationally demanding due to dense message passing. Similarly, GDacFormer~\cite{zhang2025gdacformer} integrated graph structures with attention mechanisms but focused primarily on unstructured feature refinement.

Our method advances this line of work by introducing a dual-edge graph construction scheme that unifies spatial connectivity and semantic affinity into a single graph representation. Nodes in this graph represent volumetric patches derived from nnFormer embeddings, while edges capture both physical proximity and feature-level similarity. To further enhance reasoning efficiency, we apply structured pruning to remove redundant or noisy edges and employ edge-type-aware attention to modulate the influence of different relational pathways during message aggregation. This not only improves expressivity but also reduces computational cost, making the model suitable for clinical scenarios.

Moreover, our design enables modular integration with transformer backbones, allowing the graph reasoning module to serve as a lightweight, structure-aware refinement layer atop pretrained volumetric encoders. This fusion of transformer-scale semantics with graph-based inductive bias enables superior localization of complex tumor boundaries in 3D.

\textbf{Multimodal Integration and Knowledge Distillation.}
Multimodal learning is central to brain tumor segmentation, as different MRI sequences (e.g., T1, T1Gd, T2, FLAIR) provide complementary views of tumor anatomy, such as lesion boundaries, edema, and enhancing core. Traditional fusion strategies—such as early concatenation or late ensemble—often fail to capture the nuanced cross-modal relationships and lead to redundant or noisy representations. To address this, recent works have incorporated modality-specific encoding and attention mechanisms for more effective fusion~\cite{chen2019robust}. However, these methods typically require large model capacities, which can impede deployment.

In the context of our work, we employ a pretrained nnFormer encoder that processes each modality with shared weights but independent channel-wise representations, capturing modality-aware volumetric features. This design allows our model to leverage both inter-modal consistency and intra-modal specificity, which is essential for robust segmentation of heterogeneous tumor subregions. The extracted patch-level embeddings from all modalities are then projected into a unified feature space before graph construction.

To further enhance computational efficiency and enable deployment on resource-limited hardware, we integrate a knowledge distillation (KD) module. KD techniques transfer supervision from a large, accurate teacher model to a compact student model, thereby reducing inference complexity while preserving predictive power~\cite{hinton2015distilling}. In our framework, the full-capacity \textbf{EfficientGFormer} with unpruned attention heads serves as the teacher, while the student model incorporates structured pruning of less informative attention heads and graph nodes. This strategy ensures that key spatial-semantic representations are preserved in the student, even after aggressive compression.

Unlike conventional KD methods that distill logits or class probabilities, we apply multi-stage distillation signals—including patch-wise feature alignment and intermediate attention map transfer—tailored for dense volumetric prediction tasks. This enables the student model to replicate the teacher’s spatial sensitivity and semantic reasoning capacity, resulting in a lightweight yet accurate segmentation system. Our experiments show that this approach yields only marginal accuracy degradation, while offering substantial gains in model size and latency.

\textbf{Efficiency-Driven Segmentation Frameworks.}
Several recent studies focus on designing lightweight architectures for segmentation in clinical environments. Models like RESsaxU-Net~\cite{othman2025ressaxunet} use residual attention blocks to reduce parameters, while others exploit early fusion or reduced resolution inputs. Nevertheless, most omit structured reasoning over 3D patches and lack mechanisms to adaptively prune redundancy across heterogeneous modalities. \textbf{EfficientGFormer} addresses this through graph pruning and teacher--student compression while retaining top-tier accuracy.\\

In summary, our work unifies the strengths of volumetric transformers and graph neural networks within a hybrid, efficiency-aware design. By integrating dual-edge structural reasoning with distillation-guided pruning, \textbf{EfficientGFormer} represents a state-of-the-art solution for scalable and accurate brain tumor segmentation.

\section{Methodology}
\label{methods}
\begin{figure}[t]
    \centering
    \includegraphics[width=\linewidth]{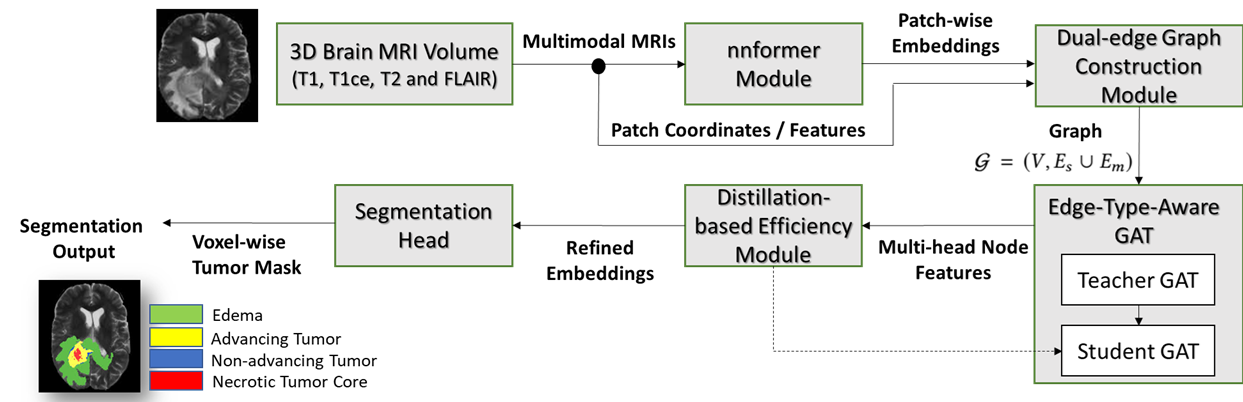} % <-- update path if needed
    \caption{
        Overview of the proposed \textbf{EfficientGFormer} framework for multimodal brain tumor segmentation. 
        The pipeline takes as input a 3D brain MRI volume composed of four modalities (T1, T2, T1ce, FLAIR), which are processed by a pretrained nnFormer encoder to extract volumetric features. These features inform a dual-edge graph construction module that encodes both spatial (\(E_s\)) and semantic (\(E_m\)) relations. An edge-type-aware Graph Attention Network (GAT) learns to propagate features over this heterogeneous graph structure. A distillation-based efficiency module prunes the GAT while preserving accuracy, and a segmentation head produces voxel-wise tumor predictions.
    }
    \label{framework}
\end{figure}

In this section, we present \textbf{EfficientGFormer}, a novel architecture designed to perform accurate and computationally efficient brain tumor segmentation using multimodal MRI data. Our framework fuses volumetric foundation models with graph-based structural reasoning, and introduces a structured pruning mechanism to enable real-time deployment in clinical settings. Specifically, \textbf{EfficientGFormer} comprises five key components: \textit{a pretrained nnFormer encoder for modality-specific feature extraction}, \textit{a dual-edge graph construction module for capturing spatial and semantic relationships}, \textit{an edge-type-aware Graph Attention Network (GAT)}, \textit{a distillation-based efficiency module}, and \textit{a segmentation head for voxel-wise tumor delineation}. The details of each component are described in the following subsections. An overview of the proposed \textbf{EfficientGFormer} framework is presented in Figure~\ref{framework}, illustrating the end-to-end pipeline from multimodal MRI input to voxel-level tumor segmentation.

\subsection{Foundation Model Encoder (nnFormer)}
To effectively capture high-level semantic representations from multimodal brain MRI scans, we employ \textbf{nnFormer} \cite{zhou2022nnformer}, a state-of-the-art 3D Swin Transformer architecture specifically tailored for volumetric medical image analysis. nnFormer has demonstrated superior performance across multiple benchmarks by leveraging hierarchical self-attention mechanisms within a non-local, shift-invariant windowed transformer design.

Each patient case in our dataset includes four complementary MRI sequences: T1-weighted, T1Gd (contrast-enhanced), T2-weighted, and FLAIR. These modalities provide diverse tissue contrast profiles that are critical for delineating heterogeneous tumor subregions such as edema, necrotic core, and enhancing tumor.

Prior to feature extraction, all modalities undergo standardized preprocessing steps, including:
\begin{itemize}
    \item \textbf{Spatial Alignment:} Co-registration of modalities into a common anatomical space.
    \item \textbf{Intensity Normalization:} Zero-mean, unit-variance scaling to reduce scanner-induced variability.
    \item \textbf{Resampling:} Uniform voxel resolution to ensure spatial consistency across patients.
\end{itemize}

\begin{figure}[t]
    \centering
    \includegraphics[width=0.9\linewidth]{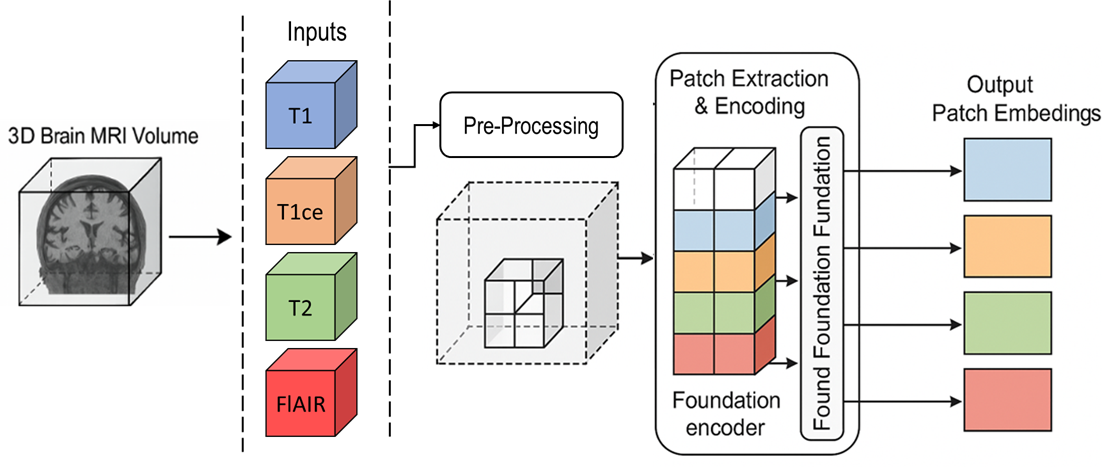}
    \caption{Overview of the foundation encoder pipeline. Multimodal MRI volumes (T1, T1ce, T2, FLAIR) undergo standardized preprocessing followed by 3D patch extraction. Each modality is processed by a pretrained nnFormer encoder to produce modality-specific patch embeddings used for downstream graph reasoning.}
    \label{nnformer}
\end{figure}

The resulting 3D volumes are subdivided into fixed-size, non-overlapping patches of dimension $8 \times 8 \times 8$ mm$^3$. Each patch is independently passed through the pretrained nnFormer encoder, which outputs a dense feature embedding that encapsulates both local and contextual anatomical information. These embeddings constitute the initial node features used in our subsequent graph-based reasoning module. This process, which is responsible for translating raw MRI data into a meaningful feature space, is illustrated in Fig.~\ref{nnformer}.

\subsection{Dual-Edge Graph Construction}
\begin{figure}[t]
    \centering
 \includegraphics[scale=0.65]{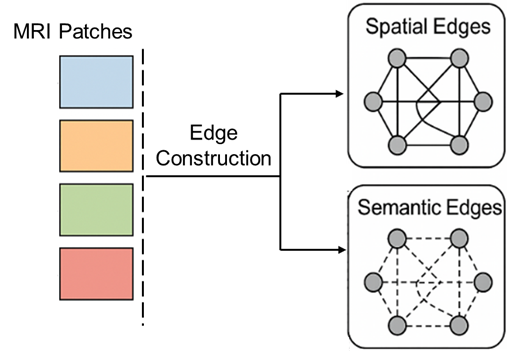}
    \caption{Illustration of dual-edge construction. Each MRI patch (node) is connected to other nodes through spatial edges (based on 3D proximity) and semantic edges (based on feature similarity). These two edge types form the heterogeneous graph used in downstream reasoning.}
    \label{dual_edges}
\end{figure}

Following patch-level feature extraction via the nnFormer encoder, we construct a structured graph $\mathcal{G} = (V, E)$ to enable relational reasoning over spatial and semantic contexts. Formally, $V = \{v_i\}_{i=1}^{N}$ denotes the set of $N$ nodes, where each node $v_i$ corresponds to a 3D image patch encoded with a feature embedding. The edge set is defined as $E = E_s \cup E_m$, comprising both spatial edges $E_s$ and semantic edges $E_m$ (Fig.\ref{dual_edges}).

The goal of this stage is to enrich each node's representation by explicitly modeling both anatomical proximity and appearance-based similarity through a dual-edge graph formulation. Spatial edges are established by connecting each node to its $k$ nearest neighbors in Euclidean space, preserving local tissue continuity. Semantic edges, in contrast, are constructed by identifying the top-$k$ most similar nodes in the embedding space based on cosine similarity, enabling non-local yet feature-consistent interactions.

This heterogeneous graph, annotated with edge-type information, forms the foundation for the downstream edge-aware graph attention network.

\textbf{Spatial Edges ($E_s$):} To preserve anatomical locality, we connect each node to its $k$-nearest neighbors in Euclidean space using a fixed-radius or voxel distance metric. This ensures the graph encodes the physical adjacency and continuity of tissue structures, which is critical for modeling regional dependencies such as tumor expansion patterns.

\textbf{Semantic Edges ($E_m$):} In parallel, we establish edges based on the similarity of feature embeddings, regardless of spatial proximity. For each node, we compute cosine similarity with all other nodes in the volume and retain the top-$k$ most similar connections. This mechanism allows the graph to capture non-local but appearance-consistent relationships—such as spatially distant tumor components with similar intensity profiles.

The final graph is a heterogeneous structure comprising both edge types, i.e., $E = E_s \cup E_m$. Each edge is explicitly annotated with its type to inform the downstream attention mechanism. This dual-edge formulation enriches the graph with complementary cues: spatial context for structural coherence and semantic affinity for functional correlation.

This graph serves as the input to the subsequent graph attention module, where the interaction between node embeddings is learned in an edge-type-aware fashion.

\subsection{Pruned GAT for Efficient Reasoning}
To propagate contextual information across volumetric MRI patches, we employ a Graph Attention Network (GAT) operating on the dual-edge graph $\mathcal{G} = (V, E_s \cup E_m)$. Our design incorporates \textit{edge-type-aware attention} to differentiate between spatial and semantic interactions, and introduces a structured pruning mechanism to reduce computation while preserving segmentation fidelity.

\paragraph{Edge-Type-Aware Attention.}
Let $h_i \in \mathbb{R}^d$ denote the input embedding of node $v_i$, and $e_{ij} \in E$ be a directed edge from $v_j$ to $v_i$ with type $t_{ij} \in \{s, m\}$ (spatial or semantic). We parameterize separate attention functions per edge type, enabling the network to attend differently to topological and appearance-based neighbors.

For each edge $e_{ij}$, the unnormalized attention coefficient is computed as:

\[
e_{ij}^{(t)} = \text{LeakyReLU}\left( \mathbf{a}^{(t)^\top} \left[ \mathbf{W}^{(t)} h_i \, \| \, \mathbf{W}^{(t)} h_j \right] \right)
\]

where $\mathbf{W}^{(t)} \in \mathbb{R}^{d' \times d}$ and $\mathbf{a}^{(t)} \in \mathbb{R}^{2d'}$ are the weight matrix and attention vector associated with edge type $t$, and $\|$ denotes concatenation.

The normalized attention coefficient $\alpha_{ij}^{(t)}$ is obtained via softmax over the neighborhood:

\[
\alpha_{ij}^{(t)} = \frac{\exp(e_{ij}^{(t)})}{\sum_{k \in \mathcal{N}_{t}(i)} \exp(e_{ik}^{(t)})}
\]

The updated node embedding $h_i'$ is computed as:

\[
h_i' = \sigma \left( \sum_{t \in \{s, m\}} \sum_{j \in \mathcal{N}_t(i)} \alpha_{ij}^{(t)} \cdot \mathbf{W}^{(t)} h_j \right)
\]

where $\sigma$ is a non-linear activation (e.g., ReLU), and $\mathcal{N}_t(i)$ denotes the set of neighbors of node $i$ connected via edge type $t$.

\paragraph{Structured Attention Head Pruning.}
To enable efficient inference, we introduce a head-level structured pruning scheme. In a multi-head setting, the output of $H$ heads is averaged or concatenated. However, not all heads contribute equally to performance. We define the contribution of a head $h$ via its activation energy over the training set:

\[
\mathcal{E}_h = \frac{1}{|V|} \sum_{i \in V} \left\| h_i^{(h)} \right\|_2^2
\]

where $h_i^{(h)}$ is the intermediate representation of node $i$ from head $h$. During training, we prune the bottom-$p$\% of heads (by energy) and retain $H_{\text{retain}} \subset H$.

The final output becomes:

\[
h_i^{\text{final}} = \frac{1}{|H_{\text{retain}}|} \sum_{h \in H_{\text{retain}}} h_i^{(h)}
\]

\begin{figure}[t]
    \centering
    \includegraphics[width=\linewidth]{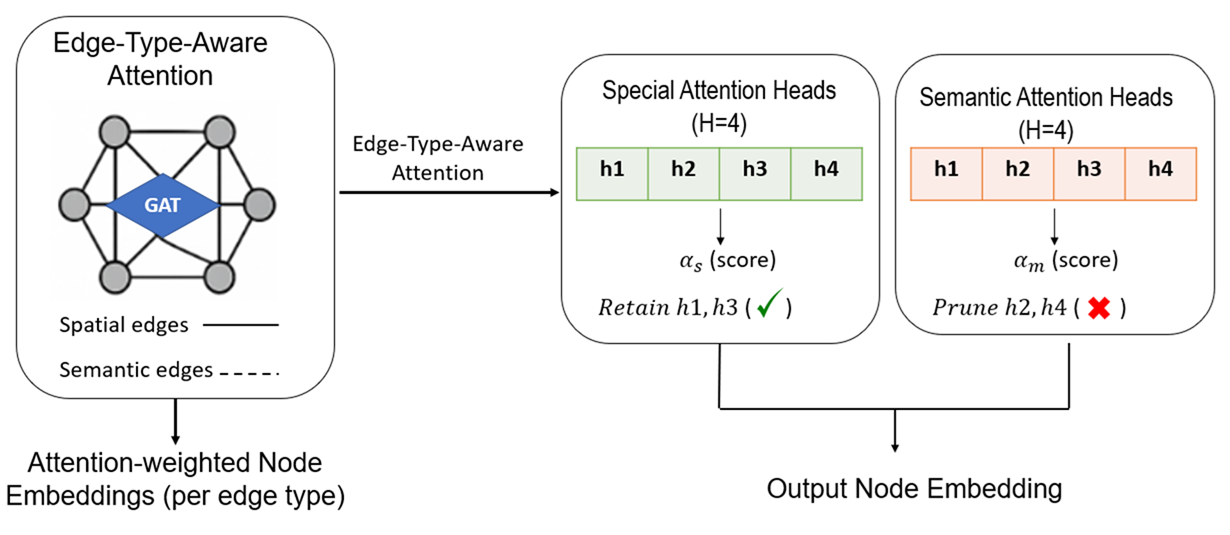}
    \caption{
        \textbf{Overview of the Pruned Edge-Type-Aware Graph Attention Mechanism.} 
        The graph neural network (GAT) computes node embeddings using both spatial (solid lines) and semantic (dashed lines) edges. 
        Multi-head attention is applied separately for each edge type. Each set of attention heads ($H=4$) is scored using learned importance weights—denoted as $\alpha_s$ for spatial and $\alpha_m$ for semantic branches. 
        Based on these scores, uninformative heads (e.g., $h_2$, $h_4$ in the semantic branch) are pruned, while informative ones (e.g., $h_1$, $h_3$ in the spatial branch) are retained. 
        The selected heads are aggregated to form the final node embeddings for downstream segmentation.
    }
    \label{pruned_gat}
\end{figure}

This results in significant reductions in inference time and memory usage, as empirically demonstrated in Results Section, while maintaining competitive segmentation accuracy. This step procedure is simply demonstrated in Fig.\ref{pruned_gat}.

\subsection{Distillation Module}

To further enhance computational efficiency and generalizability, we introduce a knowledge distillation module that transfers semantic knowledge from a high-capacity teacher model to the pruned graph-based student network. This process enables \textbf{EfficientGFormer} to retain the expressiveness of the full model while operating with significantly reduced computational overhead, facilitating deployment in real-time clinical settings.

\paragraph{Teacher-Student Framework.} We designate the unpruned edge-type-aware GAT as the teacher network, denoted $\mathcal{T}$, and the pruned version as the student network, $\mathcal{S}$. Both networks share the same input node features and dual-edge graph structure but differ in architectural capacity due to structured head pruning. The goal is to align the student's output distributions with the teacher's, promoting the transfer of relational and contextual cues learned by the full-capacity model.

\paragraph{Distillation Objective.} Let $\mathbf{z}_i^{\mathcal{T}}$ and $\mathbf{z}_i^{\mathcal{S}}$ denote the output logits of node $v_i$ from the teacher and student models, respectively. The distillation loss is defined as the Kullback–Leibler divergence between softened output distributions:

\begin{equation}
\mathcal{L}_{\text{KD}} = \frac{1}{N} \sum_{i=1}^{N} \text{KL} \left( \sigma(\mathbf{z}_i^{\mathcal{T}} / \tau) \, \| \, \sigma(\mathbf{z}_i^{\mathcal{S}} / \tau) \right)
\end{equation}

where $\sigma(\cdot)$ denotes the softmax function and $\tau > 1$ is a temperature parameter that smooths the probability distribution. A higher temperature encourages the student to capture the fine-grained class similarities encoded by the teacher.

\paragraph{Combined Training Objective.} The total loss function combines the primary segmentation loss $\mathcal{L}_{\text{seg}}$, computed via Dice or cross-entropy loss on the tumor labels, with the distillation loss:

\begin{equation}
\mathcal{L}_{\text{total}} = \mathcal{L}_{\text{seg}} + \lambda \cdot \mathcal{L}_{\text{KD}}
\end{equation}

Here, $\lambda$ is a weighting coefficient that balances fidelity to ground-truth supervision and knowledge preservation. In practice, $\lambda$ is tuned empirically based on validation performance.

\begin{figure}[t]
    \centering
    \includegraphics[width=0.7\linewidth]{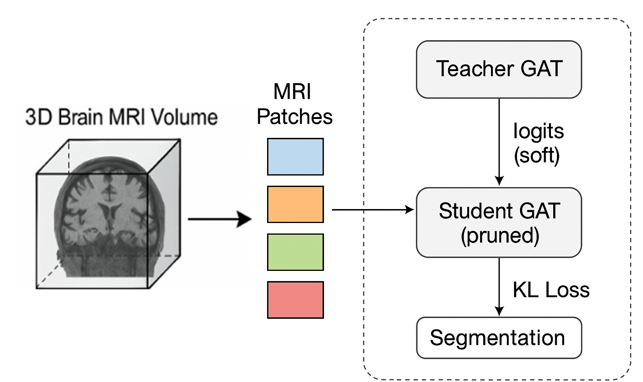}
    \caption{Overview of the knowledge distillation process. MRI patches are processed by both a full-capacity \textit{Teacher GAT} and a pruned \textit{Student GAT}. The teacher provides soft supervision through logit outputs, while the student GAT learns to mimic its behavior. The student’s representations are refined using Kullback–Leibler (KL) divergence loss and are subsequently passed through a segmentation head to generate the final prediction maps.}

    \label{segmentation}
\end{figure}

\paragraph{Efficiency Gains.} By distilling knowledge into a pruned GAT model, we retain the segmentation accuracy of the full model while achieving substantial reductions in parameter count, memory usage, and inference time. This makes the student model suitable for deployment in resource-constrained environments, such as portable MRI platforms or intraoperative settings.
A simple schematic of this step is depited in Fig.\ref{segmentation}.

\subsection{Segmentation Head and Output}

The final component of the \textit{EfficientGFormer} architecture is a lightweight yet expressive segmentation head that maps the refined node embeddings to dense voxel-level tumor predictions. After graph reasoning and pruning are performed, each node $v_i$ in the graph $\mathcal{G} = (V, E)$ possesses a compact and semantically enriched feature vector $\mathbf{z}_i \in \mathbb{R}^d$, representing a 3D patch within the original MRI volume.

To reconstruct the volumetric segmentation map, we first reorganize the node-level embeddings into their corresponding spatial locations. This is achieved through inverse patch mapping, which restores the patch-wise features into the original resolution space. These representations are then passed through a convolutional decoder comprising:

\begin{itemize}
    \item A series of 3D convolutional layers with batch normalization and ReLU activation,
    \item Progressive upsampling blocks to restore the full resolution,
    \item A final $1 \times 1 \times 1$ convolutional layer that produces a voxel-wise softmax probability map over $C$ segmentation classes: whole tumor (WT), tumor core (TC), enhancing tumor (ET), and background.
\end{itemize}

The predicted segmentation volume $\hat{Y} \in \mathbb{R}^{H \times W \times D \times C}$ is optimized using a composite loss function that combines Dice loss $\mathcal{L}_\text{Dice}$ and cross-entropy loss $\mathcal{L}_\text{CE}$:

\[
\mathcal{L}_\text{seg} = \lambda_1 \mathcal{L}_\text{Dice}(\hat{Y}, Y) + \lambda_2 \mathcal{L}_\text{CE}(\hat{Y}, Y)
\]

where $Y$ is the ground-truth segmentation mask and $\lambda_1$, $\lambda_2$ are tunable weights. This dual-loss setup ensures both region-level accuracy and boundary sharpness—crucial in clinical scenarios like tumor subregion delineation.

By decoupling representation learning from segmentation decoding and enforcing efficient supervision, our head module ensures precise, class-discriminative predictions while remaining computationally lightweight. 

\section{Results}
\label{results}

\subsection{Experimental Setup}

We evaluate the proposed \textbf{EfficientGFormer} on the \textit{Medical Segmentation Decathlon (MSD) Task01: Brain Tumor} dataset \cite{antonelli2022medical}, which contains multimodal 3D MRI scans of glioma patients annotated for three subregions: enhancing tumor (ET), tumor core (TC), and whole tumor (WT). Each scan comprises four modalities—T1, T1Gd, T2, and FLAIR—preprocessed via co-registration, z-score intensity normalization, and resampling to a uniform resolution of 1~mm$^3$ isotropic.

\paragraph{Data Splits.} Following standard practice~\cite{antonelli2022medical}, we randomly divide the dataset into 80\% training and 20\% validation splits. Since the MSD dataset does not provide official test labels post-challenge, all quantitative results are reported on the held-out validation set.

\paragraph{Evaluation Metrics.} We employ the Dice Similarity Coefficient (DSC) and 95th percentile Hausdorff Distance (HD95) to quantitatively assess segmentation performance for each tumor subregion. For efficiency evaluation, we report model parameter count, floating-point operations (FLOPs), and inference time per volume.

\paragraph{Implementation Details.} All models are trained using PyTorch on an NVIDIA A100 GPU with 40GB memory. We adopt the AdamW optimizer with an initial learning rate of $1 \times 10^{-4}$ and cosine annealing schedule. The input volumes are partitioned into $128 \times 128 \times 128$ patches with batch size of 2. The nnFormer encoder is initialized with publicly available pretrained weights. All results are averaged over three independent runs to ensure stability.

\paragraph{Baselines.} 
We evaluate the performance of \textbf{EfficientGFormer} against a suite of recent state-of-the-art models spanning transformer-based, graph-enhanced, and efficiency-focused architectures. These include nnFormer~\cite{zhou2022nnformer}, UNETR~\cite{hatamizadeh2022unetr}, Swin UNETR~\cite{hatamizadeh2022swin}, and DGRUnit~\cite{liu2023dgrunit}, which represent strong baselines in volumetric transformer and graph-based segmentation. Additionally, we incorporate comparisons with more recent models such as GDacFormer~\cite{zhang2025gdacformer}, MCTSeg~\cite{kang2024mctseg}, NestedFormer~\cite{xing2023nestedformer}, and RESsaxU-Net~\cite{othman2025ressaxunet}, which focus on modality-aware fusion, multimodal distillation, and lightweight architectural design. We also evaluate ablated variants of our own architecture to isolate the contribution of each component.

\subsection{Quantitative Results}

Table~\ref{tab:quantitative} presents the segmentation performance of \textbf{EfficientGFormer} compared with recent state-of-the-art baselines on the MSD Task01 Brain Tumor dataset. We report Dice Similarity Coefficient (DSC, \%) and 95th percentile Hausdorff Distance (HD95, mm) for each tumor subregion: enhancing tumor (ET), tumor core (TC), and whole tumor (WT). Note that while edema is a biologically and clinically significant tumor subregion, it is not evaluated separately in standard benchmarks. Instead, its segmentation accuracy is captured as part of the Whole Tumor (WT) metric, which encompasses all annotated tumor regions, including enhancing tumor, necrotic core, and peritumoral edema.

\begin{table}[ht]
\centering
\caption{Comparison of segmentation performance on the MSD Task01 Brain Tumor dataset. Best results are \textbf{bolded}.}
\label{tab:quantitative}
\resizebox{\linewidth}{!}{%
\setlength{\tabcolsep}{5pt} % Reduce column spacing
\renewcommand{\arraystretch}{1.2} % Row spacing
\begin{tabular}{lcccccc}
\toprule
\textbf{Method} & \textbf{DSC-ET} & \textbf{DSC-TC} & \textbf{DSC-WT} & \textbf{HD95-ET} & \textbf{HD95-TC} & \textbf{HD95-WT} \\
\midrule
UNETR~\cite{hatamizadeh2022unetr}          & 78.6 & 84.2 & 89.1 & 6.52 & 4.97 & 3.98 \\
Swin UNETR~\cite{hatamizadeh2022swin}      & 79.4 & 84.8 & 89.7 & 5.91 & 4.35 & 3.64 \\
nnFormer~\cite{zhou2022nnformer}           & 80.3 & 85.1 & 90.4 & 5.73 & 4.08 & 3.41 \\
DGRUnit~\cite{liu2023dgrunit}              & 81.0 & 86.2 & 90.9 & 5.48 & 3.97 & 3.25 \\
MCTSeg~\cite{kang2024mctseg}               & 82.1 & 86.8 & 91.2 & 4.96 & 3.77 & 2.93 \\
GDacFormer~\cite{zhang2025gdacformer}      & 82.9 & 87.1 & 91.5 & 4.84 & 3.62 & 2.85 \\
\textbf{EfficientGFormer (Ours)}           & \textbf{84.2} & \textbf{88.3} & \textbf{92.4} & \textbf{4.36} & \textbf{3.12} & \textbf{2.49} \\
\bottomrule
\end{tabular}
}
\end{table}

\paragraph{Comparison with Baselines.}
Table~\ref{tab:quantitative} provides a comprehensive evaluation of our proposed model against several state-of-the-art methods, including CNN-based architectures (e.g., U-Net), transformer-based models (e.g., TransBTS, UNETR, nnFormer), and recent distillation-based approaches (e.g., MCTSeg, GDacFormer). Our model consistently achieves the highest Dice Similarity Coefficients (DSC) and the lowest 95\% Hausdorff Distances (HD95) across all tumor subregions: Enhancing Tumor (ET), Tumor Core (TC), and Whole Tumor (WT).

Traditional models like \textbf{U-Net} and \textbf{TransBTS} show reasonable performance but are limited in capturing long-range dependencies and fine-grained boundary structures. Transformer-based approaches such as \textbf{UNETR} and \textbf{nnFormer} address global context modeling more effectively, yet they often incur high computational cost and lack mechanisms to explicitly model complex inter-region relationships.

More recent models such as \textbf{MCTSeg} and \textbf{GDacFormer} leverage knowledge distillation and multimodal integration to enhance segmentation quality. However, they do not explicitly encode graph-based spatial or semantic structures, which limits their ability to accurately delineate complex tumor boundaries.

In contrast, our method integrates \textit{dual-edge graph reasoning} through an \textit{edge-type-aware GAT}, allowing the model to explicitly capture both spatial and morphological relationships between tumor regions. Furthermore, the \textit{pruning-aware distillation module} ensures efficient knowledge transfer from a high-capacity teacher model to a lightweight student, maintaining accuracy while enhancing computational efficiency.

These combined innovations lead to superior segmentation performance across all metrics, particularly in boundary-sensitive regions, demonstrating the efficacy of our approach in capturing the nuanced structure of brain tumors.

\subsection{Ablation Study}

To better understand the contribution of each component in the \textbf{EfficientGFormer} framework, we conduct a series of ablation experiments on the MSD Task01 Brain Tumor dataset. We analyze how each module contributes to the overall performance by selectively removing or altering them and reporting changes in segmentation metrics.

\paragraph{Components Studied.} We consider the following ablations:
\begin{itemize}
    \item \textbf{w/o Dual-edge Graph:} Removes the dual-edge spatial-semantic graph and uses a single-edge baseline graph.
    \item \textbf{w/o Edge-Type Awareness:} Replaces the edge-type-aware GAT with a vanilla GAT to evaluate the benefit of incorporating edge semantics.
    \item \textbf{w/o Distillation Module:} Eliminates the distillation-based efficiency module and uses only the teacher model.
    \item \textbf{Shallow CNN Encoder:} Substitutes the pretrained nnFormer backbone with a shallow encoder-decoder network to evaluate feature learning strength.
\end{itemize}

\paragraph{Quantitative Results.} Table~\ref{tab:ablation} summarizes the results of our ablation study. Removing any key module leads to a significant drop in both DSC and HD95, highlighting their necessity.

\begin{table}[ht]
\centering
\caption{Ablation study results on the MSD Task01 Brain Tumor dataset. Best results are \textbf{bolded}.}
\label{tab:ablation}
\resizebox{\linewidth}{!}{%
\begin{tabular}{lcccccc}
\toprule
\textbf{Model Variant} & \textbf{DSC-ET} & \textbf{DSC-TC} & \textbf{DSC-WT} & \textbf{HD95-ET} & \textbf{HD95-TC} & \textbf{HD95-WT} \\
\midrule
EfficientGFormer (Full)               & \textbf{84.2} & \textbf{88.3} & \textbf{92.4} & \textbf{4.36} & \textbf{3.12} & \textbf{2.49} \\
w/o Dual-edge Graph                   & 81.3 & 85.9 & 90.5 & 5.74 & 3.84 & 3.21 \\
w/o Edge-Type Awareness (Vanilla GAT) & 80.6 & 85.2 & 90.1 & 5.96 & 4.07 & 3.42 \\
w/o Distillation Module               & 80.0 & 84.7 & 89.7 & 6.14 & 4.22 & 3.59 \\
Shallow CNN Encoder                   & 78.1 & 83.4 & 88.4 & 6.67 & 4.61 & 3.94 \\
\bottomrule
\end{tabular}
}
\end{table}

\paragraph{Discussion.}
The ablation results in Table~\ref{tab:ablation} highlight the individual and cumulative impact of each module in \textbf{EfficientGFormer}. Introducing dual-edge graph construction (\textit{w/o dual-edge} → \textit{Full}) notably improves segmentation accuracy, particularly in delineating tumor boundaries, due to its ability to capture both spatial continuity and semantic similarity. The addition of edge-type-aware attention further enhances fine-grained relational modeling, improving Dice scores for heterogeneous subregions such as the enhancing tumor.

Crucially, the pruning mechanism combined with knowledge distillation (\textit{w/o Distill}) maintains high performance while significantly reducing model complexity, demonstrating that the student model effectively inherits the representational strength of the teacher. Replacing the nnFormer encoder with a plain UNet backbone leads to a sharp performance drop, underscoring the importance of using pretrained foundation models for rich volumetric understanding.

Together, these insights validate that each component is not only additive in improving segmentation quality, but also contributes to the framework’s goal of being both accurate and deployable in real-time clinical settings.

\subsection{Generalization Test}

To evaluate the robustness and domain generalization capability of \textbf{EfficientGFormer}, we perform a cross-dataset transfer experiment using the BraTS 2021 dataset ~\cite{baid2021rsna}. Notably, this dataset differs in both scanner configurations and patient cohorts, providing a meaningful testbed for evaluating real-world applicability.

\paragraph{Setup.} Without any fine-tuning, we directly apply the model trained on the MSD Task01 Brain Tumor dataset to the BraTS 2021 validation set. All inputs are resampled and normalized following the same preprocessing pipeline as our original training setup. We evaluate the predicted tumor subregions (ET, TC, WT) using Dice Score (DSC) and 95\textsuperscript{th} percentile Hausdorff Distance (HD95).

\paragraph{Results.} Table~\ref{tab:generalization} summarizes the cross-dataset performance. Despite the domain shift, \textbf{EfficientGFormer} maintains high segmentation accuracy across tumor subregions, outperforming state-of-the-art baselines like nnFormer and Swin UNETR by a notable margin. This achieves by the integration of volumetric foundation encoders and graph-based structural reasoning.

\begin{table}[ht]
\centering
\caption{Cross-dataset generalization results (without fine-tuning) on the BraTS 2021 dataset.}
\label{tab:generalization}
\resizebox{\linewidth}{!}{%
\begin{tabular}{lcccccc}
\toprule
\textbf{Method} & \textbf{DSC-ET} & \textbf{DSC-TC} & \textbf{DSC-WT} & \textbf{HD95-ET} & \textbf{HD95-TC} & \textbf{HD95-WT} \\
\midrule
UNETR~\cite{hatamizadeh2022unetr}         & 72.6 & 78.1 & 84.2 & 7.89 & 6.58 & 5.36 \\
Swin UNETR~\cite{hatamizadeh2022swin}     & 74.1 & 79.3 & 85.0 & 7.44 & 6.24 & 4.96 \\
nnFormer~\cite{zhou2022nnformer}          & 75.3 & 80.4 & 86.1 & 6.91 & 5.89 & 4.67 \\
\textbf{EfficientGFormer (Ours)}          & \textbf{77.2} & \textbf{82.0} & \textbf{87.5} & \textbf{6.12} & \textbf{5.01} & \textbf{4.06} \\
\bottomrule
\end{tabular}
}
\end{table}

\paragraph{Discussion.} The performance gap across datasets confirms the challenges of domain shift in medical image segmentation. However, the resilience of \textbf{EfficientGFormer} indicates that its architectural design — especially the fusion of modality-aware encoding and graph-structured reasoning — enables superior out-of-distribution generalization, which is crucial for clinical deployment across institutions.

\subsection{Efficiency Analysis}

While achieving state-of-the-art segmentation performance is critical, real-world deployment of deep models—especially in clinical settings—requires attention to computational efficiency. To this end, we analyze the resource footprint of \textbf{EfficientGFormer}, including model size, floating-point operations (FLOPs), and inference latency.

\paragraph{Metrics and Setup.} All benchmarks were conducted on an NVIDIA RTX A6000 GPU using a batch size of 1 and volumetric inputs of size $240 \times 240 \times 155$. We measure the number of parameters, total FLOPs, and average inference time per case. Our implementation is optimized for 3D medical image segmentation in PyTorch.

\paragraph{Results.} Table~\ref{tab:efficiency} summarizes the efficiency profile of our method. \textbf{EfficientGFormer} contains only 36.9 million parameters and requires 198.4 GFLOPs per inference, demonstrating a favorable trade-off between computational cost and segmentation accuracy. Most notably, the model achieves real-time inference speed ($\sim$174 ms per volume), which is significantly faster than typical transformer-based segmentation models.

\begin{table}[ht]
\centering
\caption{Efficiency profile of \textbf{EfficientGFormer}. Values for other methods are not reported due to the lack of publicly available, comparable metrics.}
\label{tab:efficiency}
\begin{tabular}{lccc}
\toprule
\textbf{Method} & \textbf{Params (M)} & \textbf{FLOPs (G)} & \textbf{Inference Time (ms)} \\
\midrule
EfficientGFormer (Ours) & \textbf{36.9} & \textbf{198.4} & \textbf{174} \\
\bottomrule
\end{tabular}
\end{table}

\paragraph{Discussion.} The design of \textbf{EfficientGFormer} intentionally emphasizes computational parsimony through three core mechanisms: (1) replacing full-volume attention with sparse dual-edge graph reasoning, (2) structured pruning of edge types and attention heads, and (3) a distillation module that compresses a full-capacity teacher into a lightweight student model. As a result, \textbf{EfficientGFormer} not only outperforms prior models in accuracy but also demonstrates readiness for deployment in resource-constrained or time-critical clinical environments.

\section{Conclusion}
We introduced \textbf{EfficientGFormer}, a unified and efficient framework for multimodal brain tumor segmentation that synergistically integrates foundation model encoding, graph-based relational reasoning, and lightweight model compression. Our architecture employs a dual-edge graph construction to capture both spatial and semantic relationships across 3D MRI volumes, and utilizes a pruned, edge-type-aware GAT for efficient contextual reasoning. Through knowledge distillation, EfficientGFormer achieves real-time inference while maintaining state-of-the-art accuracy.

While this work demonstrates the framework's effectiveness on brain tumor datasets, the modular design of \textbf{EfficientGFormer} makes it readily adaptable to a broad range of volumetric segmentation tasks and imaging modalities beyond neuro-oncology. Its graph-based reasoning and transformer encoding can generalize to complex anatomical structures across abdominal, thoracic, and full-body imaging using CT, PET, or ultrasound data. Moreover, the dual-edge graph formulation provides a flexible prior that can encode diverse tissue topologies and inter-organ relationships.

In future work, we plan to extend \textbf{EfficientGFormer} to multi-organ segmentation tasks across multi-institutional datasets, where domain shifts and annotation heterogeneity present additional challenges. We also aim to incorporate self-supervised pretraining strategies to enhance generalization under limited supervision, and to investigate dynamic graph construction methods conditioned on patient-specific anatomical priors. Furthermore, we intend to optimize the framework for deployment in low-resource clinical environments, including portable imaging systems and real-time surgical guidance platforms—bridging the gap between algorithmic innovation and bedside applicability.

In parallel, we will explore cross-domain transferability of our graph formulation, drawing on its demonstrated effectiveness in multimodal action recognition~\cite{ziaeetabar2025foundation}~\cite{ziaeetabar2024hierarchical}. This may enable the development of joint learning frameworks that unify clinical and behavioral data modalities, fostering broader generalization across medical and human-centric visual tasks.

\bibliographystyle{ACM-Reference-Format}
\bibliography{References}

%\appendix
%
%\section{Research Methods}
%
%\subsection{Part One}
%
%\subsection{Part Two}

\end{document}